\documentclass{article} 
\usepackage{iclr2020_conference,times}

\usepackage{booktabs}
\usepackage{multirow}
\usepackage{soul}
\usepackage{color}
\usepackage{hyperref}
\usepackage{url}
\usepackage{arydshln}
\usepackage{amsmath,amsfonts,amssymb,bm}
\usepackage{enumitem}
\usepackage{caption}
\usepackage{subcaption}
\usepackage{graphicx}

\usepackage{amsmath,amsfonts,bm}

\def\eqref#1{equation~\ref{#1}}

\def\1{\bm{1}}

\DeclareMathAlphabet{\mathsfit}{\encodingdefault}{\sfdefault}{m}{sl}
\SetMathAlphabet{\mathsfit}{bold}{\encodingdefault}{\sfdefault}{bx}{n}

\usepackage{hyperref}
\usepackage{url}

\definecolor{mydarkblue}{rgb}{0,0.08,0.45}
\hypersetup{
 colorlinks=true,
 citecolor=mydarkblue,
 urlcolor=mydarkblue
 }

\title{DC-BERT: Decoupling Question and Document \\ for Efficient Contextual Encoding}

\author{
Yuyu Zhang\thanks{Equal contribution}$^{\ \,,1}$, Ping Nie$^{*,2}$, Xiubo Geng$^3$, Arun Ramamurthy$^4$, Le Song$^{1}$ \& Daxin Jiang$^{3}$ \\
$^1$Georgia Institute of Technology \ \ $^2$Peking University \\
$^3$Microsoft \ \ $^4$Siemens Corporate Technology \\
\texttt{\{yuyu,lsong\}@gatech.edu,ping.nie@pku.edu.cn} \\
\texttt{\{xigeng,djiang\}@microsoft.com,arun.ramamurthy@siemens.com}\\
}

\iclrfinalcopy 

\begin{document}

\maketitle

\begin{abstract}
Recent studies on open-domain question answering have achieved prominent performance improvement using pre-trained language models such as BERT. State-of-the-art approaches typically follow the ``retrieve and read'' pipeline and employ BERT-based reranker to filter retrieved documents before feeding them into the reader module. The BERT retriever takes as input the concatenation of question and each retrieved document. Despite the success of these approaches in terms of QA accuracy, due to the concatenation, they can barely handle high-throughput of incoming questions each with a large collection of retrieved documents. To address the efficiency problem, we propose DC-BERT, a decoupled contextual encoding framework that has dual BERT models: an \textit{online} BERT which encodes the question only once, and an \textit{offline} BERT which pre-encodes all the documents and caches their encodings. On SQuAD Open and Natural Questions Open datasets, DC-BERT achieves 10x speedup on document retrieval, while retaining most (about 98\%) of the QA performance compared to state-of-the-art approaches for open-domain question answering.
\end{abstract}

\section{Introduction}



Open-domain question answering (QA) is an important and challenging task in natural language processing, which requires a machine to find the answer by referring to large unstructured text corpora without knowing which documents may contain the answer. Recently, pre-trained language models such as BERT~\citep{devlin2019bert} have boosted up the performance of open-domain QA on several benchmark datasets, such as HotpotQA~\citep{yang-etal-2018-hotpotqa} and Natural Questions~\citep{kwiatkowski2019natural}. With high-quality contextual encodings, BERT-based approaches~\citep{nie-etal-2019-revealing,hu-etal-2019-retrieve} have dominated the leaderboards and significantly outperformed previous RNN and CNN based approaches~\citep{qi2019answering,jiang-bansal-2019-self,min-etal-2018-efficient,wei2018qanet,wang2018reinforced}.

Recent studies on open-domain QA typically follow the ``retrieve and read'' pipeline initiated by \citet{chen-etal-2017-reading}, which combines information retrieval (IR) and machine reading comprehension (MRC) modules as a pipeline: the former retrieves the documents using off-the-shelf IR systems based on TF-IDF or BM25, and the latter reads the retrieved documents to extract answer.
The IR systems are purely based on n-gram matching and have shallow understanding of the context. Thus, documents that contain the correct answer may not be ranked among the top by IR systems~\citep{Asai2020Learning}.
If we simply feed more documents into the reader module to increase the chance of hitting under-ranked documents that contain the answer, it can be computationally expensive and bring more noise to the reader module, making it harder to find the answer.
To alleviate this problem, state-of-the-art approaches~\citep{nie-etal-2019-revealing,hu-etal-2019-retrieve,lee-etal-2018-ranking} have proposed to train a BERT-based reranker, which is a binary classifier to filter retrieved documents before  feeding them into the reader module. The input of the BERT retriever is the concatenation of a question and each retrieved document, formulated as:
\begin{gather*}
    [CLS] \; Question \; [SEP] \; Document \; [SEP],
\end{gather*}
where $[CLS]$ is the pooling token. Due to the high-quality contextual encodings generated by BERT, such methods significantly improve the retrieval performance over non-parameterized IR systems, and thus boost up the answer accuracy for open-domain QA. However, due to the concatenation, these approaches have to repeatedly encode a question with each of the retrieved document, which are hard to handle high-throughput incoming questions each with a large collection of retrieved documents. This severe efficiency problem prohibits existing approaches for open-domain QA from being deployed as real-time QA systems.

\begin{figure}
    \centering
    \includegraphics[width=.8\textwidth]{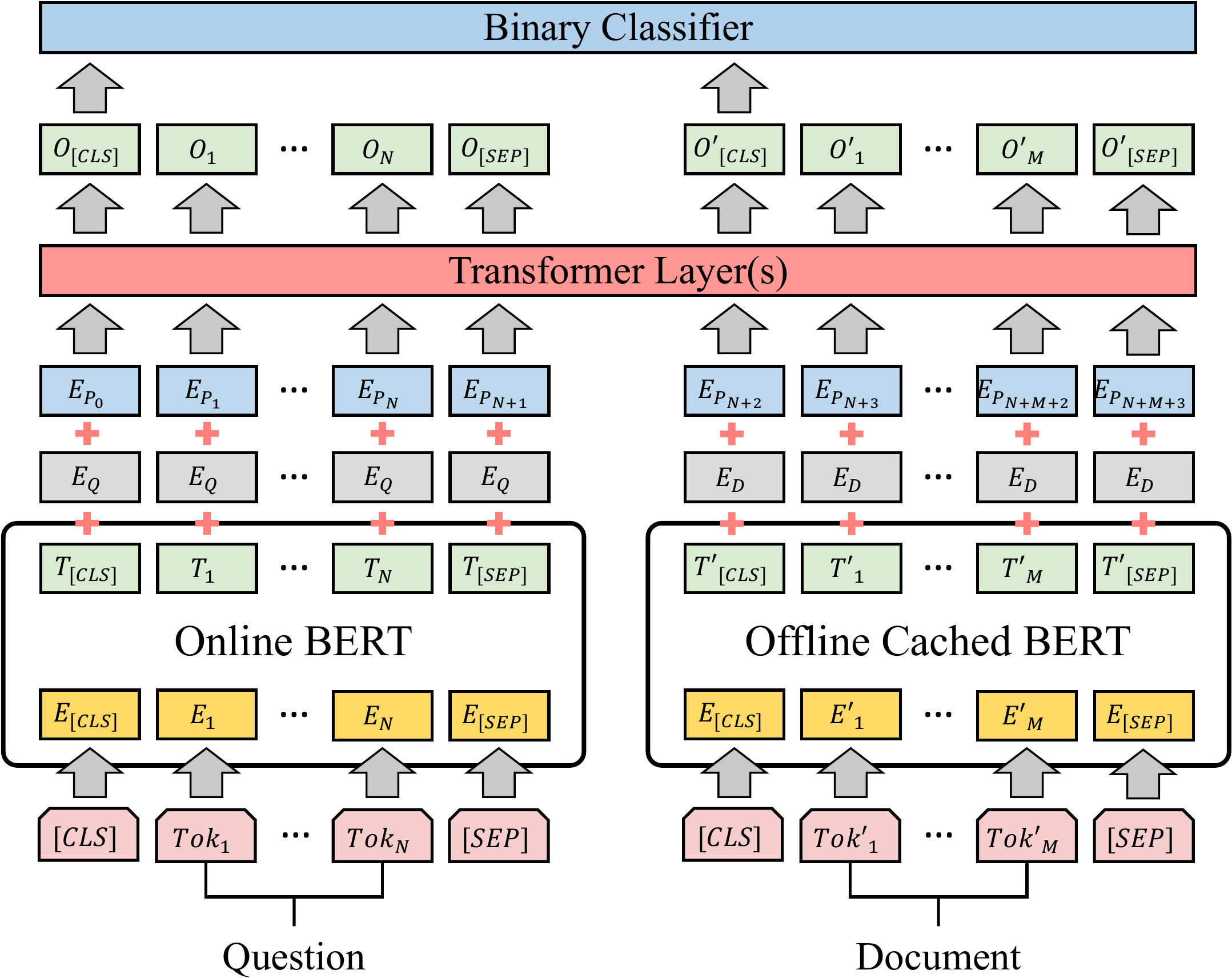}
    \caption{Overview architecture of DC-BERT, which decouples question and document for efficient contextual encoding. The binary classifier predicts whether the document is relevant to the question.}
    \label{fig:overview}
\end{figure}

Recent studies \citep{tenney-etal-2019-bert,hao-etal-2019-visualizing} have probed and visualized BERT to understand its effectiveness, which show that the lower layers of BERT encode more local syntax information such as part-of-speech (POS) tags and constituents, while the higher layers tend to capture more complex semantics relying on wider contexts. Inspired by these observations, we propose DC-BERT, which decouples the lower layers of BERT into local contexts (question and document), and then applies Transformer layers on top of the independent encodings to enable question-document interactions. As illustrated in Figure~\ref{fig:overview}, DC-BERT has two separate BERT models: an \textit{online} BERT which encodes the question only once, and an \textit{offline} BERT which pre-encodes all the documents and caches their encodings. With caching enabled, DC-BERT can instantly read out the encoding of any document. The decoupled encodings of question and document are then fed into the Transformer layers with global position and type embeddings for question-document interactions, which produces the contextual encodings of the (question, document) pair. DC-BERT can be applied to both document retriever and reader. In this work, we focus on speeding up the retriever, since the number of documents retrieved per question can be fairly large, while the number of documents fed into the reader module is much controlled. Therefore, it is more important to address the efficiency problem of the document retriever.

Speeding up BERT-based models for efficient inference is an active field of research. Previous works related to this direction mainly include: 1) model compression, where methods are proposed to reduce the model size through weight pruning or model quantization~\citep{jacob2018quantization}; and 2) model distillation, where methods are proposed to train a small student network from a large teacher network, such as DistilBERT~\citep{sanh2019distilbert}. Both lines of research are actually orthogonal to our work, since the compressed / distilled BERT model can be combined with DC-BERT. In the experiments, we also compare with quantized BERT and DistilBERT, showing that DC-BERT has significant performance advantage over these methods.

Our main contributions are summarized as follows:
\begin{itemize}[leftmargin=*,nolistsep,nosep]
    \item \textit{Decoupled QA encoding}: We propose to decouple question and document for efficient contextual encoding. To the best of our knowledge, our work is the first to explore a combination of local- and global-context encoding with BERT for open-domain QA. 
    \item \textit{Effective question-document interactions}: We propose an effective model architecture for question-document interactions, which employs trainable global embeddings with Transformer layers.
    \item \textit{Fast document retrieval}: We successfully apply DC-BERT to document reranking, a key component in open-domain QA, by making it over 10x faster than the existing approaches, while retaining most (about 98\%) of the QA performance on benchmark datasets.
    \item \textit{New evaluation metrics}: We propose two symmetric new evaluation metrics to gauge the retriever's capability of discovering documents that have low TF-IDF scores but contain the answer.
\end{itemize}

\section{Methodology}

The overall architecture of DC-BERT (Figure~\ref{fig:overview}) consists of a dual-BERT component for decoupled encoding, a Transformer component for question-document interactions, and a classifier component for document reranking.


\textbf{Dual-BERT component.} DC-BERT contains two BERT models to independently encode the question and each retrieved document. During training, the parameters of both BERT models are updated to optimize the learning objective, which is described later in the classifier component part. Once the model is trained, we pre-encode all the documents and store their encodings in an \textit{offline} cache. During testing, we encode the question only once using the \textit{online} BERT, and instantly read out the cached encodings of all the candidate documents retrieved by an off-the-shelf IR system.
Compared to the existing ``concatenate and encode'' approaches that concatenate the question and each retrieved document, DC-BERT only encodes a question for once, which reduces the computational cost of BERT lower layers from $O(N_q N_d (L_q+L_d)^2)$ to $O(N_q (L_q^2 + N_d L_d^2))$ where $N_q$ denote the number of questions, $N_d$ denote the number of retrieved documents for each question, $L_q$ and $L_d$ denote the average number of tokens of each question and document, respectively. Moreover, the decoupled BERT also enables caching of the document encodings offline, which further reduces the computational cost to $O(N_q L_q^2)$.

In fact, a number of previous RNN and CNN based QA approaches, such as BiDAF~\citep{seo2017bidirectional}, DCN~\citep{xiong2017dynamic} and QANet~\citep{wei2018qanet}, also decouple the encoding of question and document, but their performance is dominated by recent BERT-based approaches. With self-attention Transformer layers, BERT is designed to perform contextual encoding by incorporating wide context from pre-training to fine-tuning. This explains why most state-of-the-art approaches for open-domain QA concatenate question and document for wide-context encoding and achieve prominent performance. Recent work \citep{lee-etal-2019-latent} attempts to independently encode question and document with BERT, and compute the inner product of their encodings to retrieve documents, but performs worse than state-of-the-art approaches. We posit that the lack of interactions between question and document can significantly hurt the performance.
Therefore, our method is designed to encapsulate question-document interactions with a Transformer component, as described below.

\textbf{Transformer component.} With the dual-BERT component, we obtain the question encoding $\mathbf{T} \in \mathbb{R}^{N \times d}$ and the document encoding $\mathbf{T'} \in \mathbb{R}^{M \times d}$, where $d$ is the dimension of word embeddings, and $N$ and $M$ are the length of the question and the document, respectively. Since the document reranking task is to predict the relevance of the document for a question, we introduce a Transformer component with trainable global embeddings to model the question-document interactions.

More specifically, we have global position embeddings $\mathbf{E}_{P_i} \in \mathbb{R}^d$ to re-encode the token at position $P_i$ in the concatenated question-document encoding sequence. We also have global type embeddings $\mathbf{E}_Q \in \mathbb{R}^d$ and $\mathbf{E}_D \in \mathbb{R}^d$ to differentiate whether the encoded token is from question or document. Both the global position and type embeddings are initialized by the position and sentence embeddings from pre-trained BERT, and will be updated during the training. These additional embeddings are added on top of the question and document encodings (with the encodings of $[CLS]$ and $[SEP]$), and then fed into the Transformer layers. The number of Transformer layers $K$ is configurable to trade-off between the model capacity and efficiency. The Transformer layers are initialized by the last $K$ layers of pre-trained BERT, and are updated during the training.

\textbf{Classifier component.} After the Transformer layers, DC-BERT treats the document reranking task as a binary classification problem to predict whether the retrieved document is relevant to the question. Following previous work \citep{das2018multistep,htut-etal-2018-training,lin-etal-2018-denoising}, we employ paragraph-level distant supervision to gather labels for training the classifier, where a paragraph that contains the exact ground truth answer span is labeled as a positive example. We parameterize the binary classifier as a MLP layer on top of the Transformer layers:
\begin{align} \label{eq:proba}
    & p(Q_i, D_j) = \sigma(\text{MLP}([o_{[CLS]}; o'_{[CLS]}])),
\end{align}
where $(Q_i, D_j)$ is a pair of question and retrieved document, and $o_{[CLS]}$ and $o'_{[CLS]}$ are the Transformer output encodings of the $[CLS]$ token of the question and the document, respectively. The MLP parameters are updated by minimizing the cross-entropy loss:
\begin{align}
    \mathcal{J} = - \sum_{(Q_i, D_j)} \Big( y \log{(p)} + (1-y) \log{(1 - p) \Big)},
\end{align}
where $y = y(Q_i, D_j)$ is the distantly supervised label, and $p = p(Q_i, D_j)$ as defined in Eq.~\eqref{eq:proba}.

\section{Experiments}


\textbf{Benchmark datasets.} We evaluate DC-BERT and other baseline methods on two popular benchmark datasets: 1) SQuAD Open~\citep{chen-etal-2017-reading}, which is composed of questions from the original crowdsourced SQuAD dataset~\citep{rajpurkar2016squad}; 2) Natural Questions Open~\citep{min2019discrete}, which is composed of questions from the original Natural Questions dataset~\citep{kwiatkowski2019natural}. The questions are created from real user queries issued to Google Search engine. For all our experiments, we use the standard splits provided with the datasets, and report the performance on the development split.

\textbf{Evaluation metrics.}
To evaluate the retriever speed, we compare the wall-clock time running on a single GPU. To evaluate the retriever ranking performance, we use the following metrics: 1) P@N, which is defined in previous work~\citep{chen-etal-2017-reading} as the percentage of questions for which the answer span appears in one of the top N documents; 2) PBT@N, a new evaluation metric that we propose to gauge the semantic retrieval capability of the document reranker, which is the percentage of questions for which at least one of the top N documents that contains the answer span is \textit{not} in the TF-IDF top N documents. In other words, this new metric measures the retriever's capability beyond the TF-IDF retriever (the higher the better); 3) PTB@N, our proposed metric that is symmetric to PBT@N, which is the percentage of questions for which at least one of the top N TF-IDF retrieved documents that contains the answer span \textit{not} in the retriever's top N documents. This metric measures the retriever's capability of retaining the relevant documents returned by TF-IDF retriever (the lower the better).
To evaluate the downstream QA performance, we follow previous works~\citep{chen-etal-2017-reading,nie-etal-2019-revealing,das2018multistep} and use the standard answer exact match (EM) score.

\textbf{Implementation details.}
We use pre-trained BERT-base model \citep{devlin2019bert} for the document reranker and pre-trained BERT-wwm
(whole word masking) model for the downstream QA model. For the standard TF-IDF retrieval, we use the released retrieval data from \citet{min2019discrete} for Natural Questions Open, and use the DrQA~\citep{chen-etal-2017-reading} TF-IDF retriever to collect 80 documents for SQuAD Open. We select top 10 documents ranked by retriever to feed into the reader module. For our method, we enable caching of the document encodings.
We set $K=1$ as the number of Transformer layers for all the experiments and vary this choice in the ablation study. We use $4e-5$ as the initial learning rate. Our model is trained with Adam optimizer~\citep{kingma2014adam}. For the binary classifier, we use a two-layer MLP with the $\tanh(\cdot)$ activation function for nonlinear transformation.


\setlength\tabcolsep{2.5pt}
\begin{table*}[t]
\caption{Performance comparison on two benchmark datasets. DC-BERT uses one Transformer layer for question-document interactions. Quantized BERT is a 8bit-Integer model. DistilBERT is a compact BERT model with 2 Transformer layers.}
\label{table:main_results}
\centering
\resizebox{\textwidth}{!}{
\begin{tabular}{@{}lccccccccc@{}}
\toprule
\multirow{3}{*}{Retriever Model} & \multicolumn{4}{c}{SQuAD Open} &  & \multicolumn{4}{c}{Natural Questions Open} \\ \cmidrule(lr){2-5} \cmidrule(l){7-10} 
 & Retriever & Retriever & Answer & EM &  & Retriever & Retriever & Answer & EM \\
 & P@10 & Speedup & EM & Drop (\%) &  & P@10 & Speedup & EM & Drop (\%) \\ \cmidrule(r){1-5} \cmidrule(l){7-10} 
BERT-base & 71.5 & 1.0x & 40.1 & 0.0 &  & 65.0 & 1.0x & 28.0 & 0.0 \\
Quantized BERT & 68.0 & 1.1x  & 39.5 & 1.5 &  & 64.3 & 1.1x & 27.5 & 1.8 \\
DistilBERT & 56.4 & 5.7x & 34.6 & 13.7 &  & 60.6 & 5.7x & 25.1 & 9.7 \\
DC-BERT & 70.1 & 10.3x & 39.2 & 2.1 &  & 63.5 & 10.3x & 27.4 & 2.0 \\ \bottomrule
\end{tabular}
}
\end{table*}

\setlength\tabcolsep{5pt}
\begin{table}[t]
\caption{Retriever performance in PBT@10 and PTB@10.}
\label{table:retrieval_pbt_ptb}
\centering
\begin{tabular}{@{}lccccc@{}}
\toprule
\multirow{2}{*}{Retriever Model} & \multicolumn{2}{c}{SQuAD} &  & \multicolumn{2}{c}{Natural Questions} \\ \cmidrule(lr){2-3} \cmidrule(l){5-6} 
 & PBT@10 & PTB@10 &  & PBT@10 & PTB@10 \\ \cmidrule(r){1-3} \cmidrule(l){5-6} 
BERT-base~\citep{devlin2019bert} & 19.8 &2.1 &  & 14.6 & 4.0 \\
Quantized BERT~\citep{jacob2018quantization} & 18.4 & 3.4 & & 14.1 & 4.5 \\
DistilBERT~\citep{sanh2019distilbert} & 14.3 & 10.2 &  & 11.5 & 7.1 \\
DC-BERT & 18.1 & 5.7 &  & 13.8 & 6.8 \\ \bottomrule
\end{tabular}
\end{table}

\setlength\tabcolsep{10pt}
\begin{table}[ht]
\caption{Ablation study results on Natural Questions Open.}
\label{table:ablation}
\centering
\begin{tabular}{@{}lccc@{}}
\toprule
\multirow{2}{*}{Retriever Model} & Retriever & Retriever & Answer \\
 & P@10 & Speedup & EM \\ \midrule
DC-BERT-Linear & 57.3 & 43.6x & 24.8 \\
DC-BERT-LSTM & 61.5 & 8.2x & 26.5 \\ 
DC-BERT & 63.5 & 10.3x & 27.4 \\ \bottomrule
\end{tabular}
\end{table}

\textbf{Baseline methods.} 1) BERT-base: a well-trained reranker using the BERT-base model~\citep{devlin2019bert}, which is a very strong baseline and represents state-of-the-art performance on both datasets. Recent work~\citep{Asai2020Learning} uses external data such as hyperlinks and data augmentation to further improve the performance, which is out of the scope of this paper; 2) Quantized BERT: a recent work~\citep{zafrir2019q8bert} that employs quantization techniques~\citep{jacob2018quantization} to compress BERT-base into a 8bit-Integer model. We use the official open-sourced code for experiments; 3) DistilBERT: a recent approach~\citep{sanh2019distilbert} that leverages knowledge distillation techniques~\citep{hinton2015distilling} to train a smaller and compact student BERT model to reproduce the behavior of the teacher BERT-base model. We use the official open-sourced code for experiments. To achieve decent speedup, we set 2 Transformer layers for the student BERT model.

\textbf{Retriever speed.} The main experimental results are summarized in Table~\ref{table:main_results}. We first compare the retriever speed. DC-BERT achieves over 10x speedup over the BERT-base retriever, which demonstrates the efficiency of our method. Quantized BERT has the same model architecture as BERT-base, leading to the minimal speedup. DistilBERT achieves about 6x speedup with only 2 Transformer layers, while BERT-base uses 12 Transformer layers.

\textbf{Retriever ranking performance.} We evaluate the ranking metrics in terms of P@10 in Table~\ref{table:main_results}. With a 10x speedup, DC-BERT still achieves similar retrieval performance compared to BERT-base on both datasets. At the cost of little speedup, Quantized BERT also works well in ranking documents. DistilBERT performs significantly worse than BERT-base, which shows the limitation of the distilled BERT model. We also report the proposed PBT@10 and PTB@10 metrics in Table~\ref{table:retrieval_pbt_ptb}. As discussed, PBT@10 is the higher the better, and PTB@10 is the lower the better. DC-BERT and Quantized BERT achieves similar performance compared to BERT-base, while DistilBERT is inferior in both metrics.

\textbf{QA performance.} As reported in Table~\ref{table:main_results}, DC-BERT and Quantized BERT retain most of the QA performance (Answer EM) compared to BERT-base, on both SQuAD Open and Natural Questions Open datasets. Due to the inferior retrieval performance, DistilBERT also performs the worst in answer accuracy. With a 10x speedup, DC-BERT only has a performance drop of about 2\%, which demonstrates the effectiveness of our method in the downstream QA task.

\begin{figure}[h]
 \centering
 \begin{subfigure}[b]{0.49\textwidth}
     \centering
     \includegraphics[width=\textwidth]{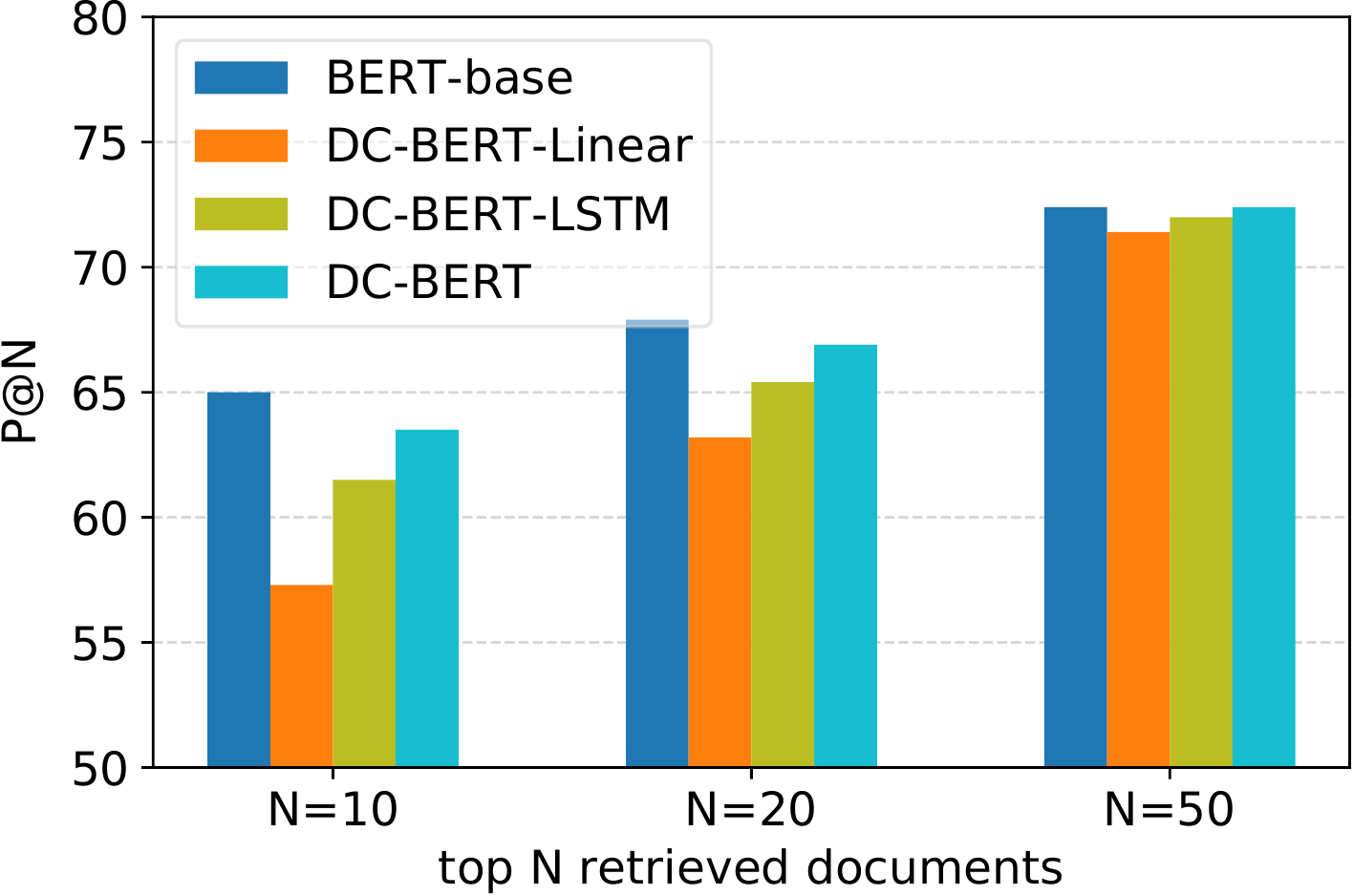}
     \caption{model architecture}
     \label{fig:reranker_p_at_n}
 \end{subfigure}
 \hfill
 \begin{subfigure}[b]{0.49\textwidth}
     \centering
     \includegraphics[width=\textwidth]{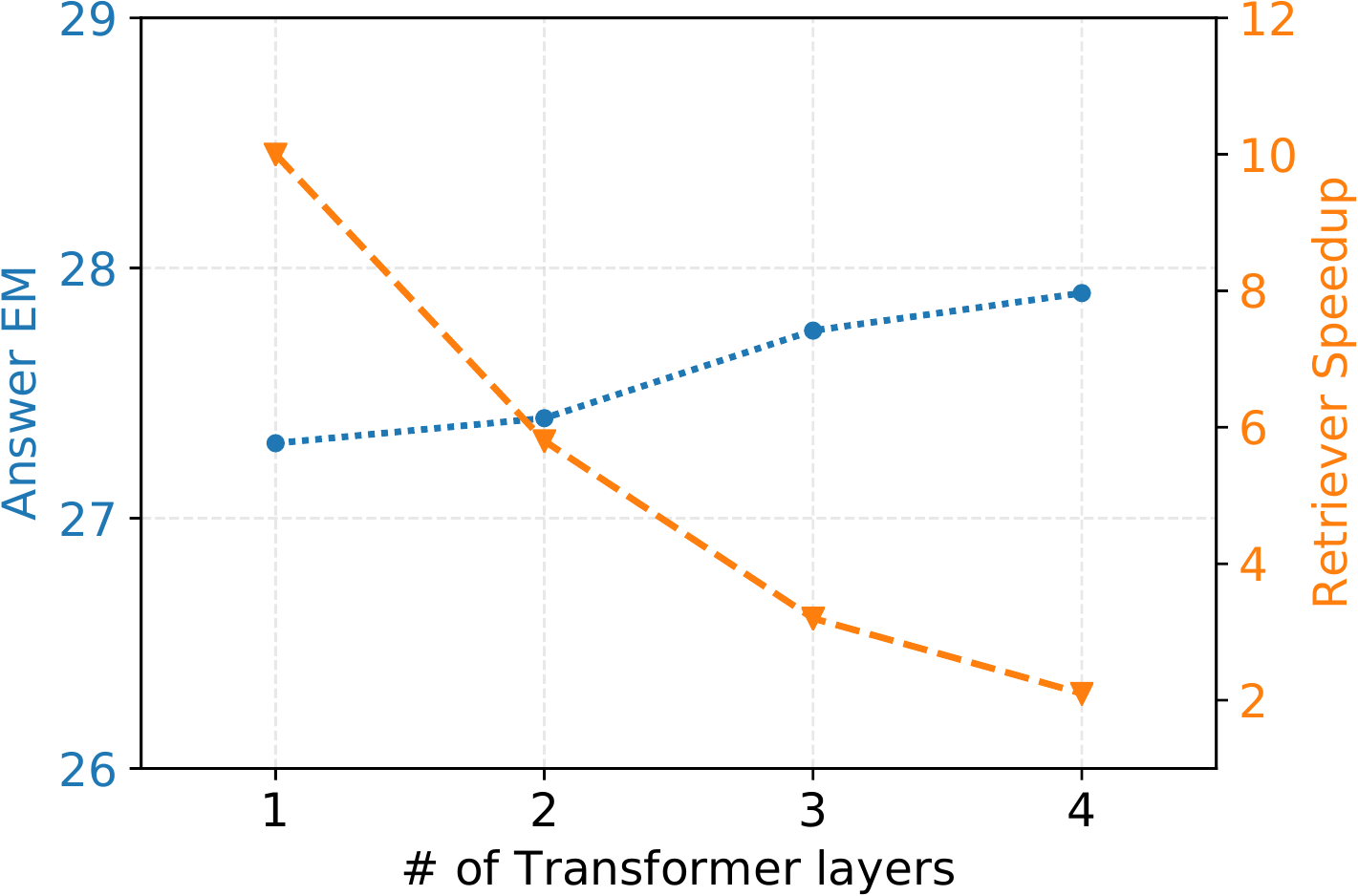}
     \caption{Transformer layers}
     \label{fig:layers_trade_off}
 \end{subfigure}
 \caption{Ablation study results on Natural Questions Open.}
 \label{fig:ablation}
\end{figure}

\textbf{Ablation study.} To further investigate the impact of our model architecture design, we compare the performance of DC-BERT and its variants, including 1) DC-BERT-Linear, which uses linear layers instead of Transformers for interaction; and 2) DC-BERT-LSTM, which uses LSTM and bilinear layers for interactions following previous work~\citep{min-etal-2018-efficient}. We report the results in Table~\ref{table:ablation}. Due to the simplistic architecture of the interaction layers, DC-BERT-Linear achieves the best speedup but has significant performance drop, while DC-BERT-LSTM achieves slightly worse performance and speedup than DC-BERT. Figure~\ref{fig:reranker_p_at_n} shows that DC-BERT consistently outperforms its variants for different number of top retrieved documents, and leads with a larger margin when retrieving less documents. We also investigate the impact of the number of Transformer layers for question-document interactions, and report the results in Figure~\ref{fig:layers_trade_off}. When we increase the number of Transformer layers, the QA performance consistently improves, and the speedup decreases due to the increased computational cost. This shows the trade-off between the model capacity and efficiency of our method.

\section{Conclusion}

This paper introduces DC-BERT to decouple question and document for efficient contextual encoding. DC-BERT has been successfully applied to document retrieval, a key component in open-domain QA, achieving 10x speedup while retaining most of the QA performance. With the capability of processing high-throughput of questions each with a large collection of retrieved documents, DC-BERT brings open-domain QA one step closer to serving real-world applications.

\clearpage

\bibliography{references_long}
\bibliographystyle{iclr2020_conference}


\end{document}